\documentclass[letterpaper, 10pt, conference]{ieeeconf}
\IEEEoverridecommandlockouts                              

\usepackage{graphics} 
\usepackage{graphicx} 
\graphicspath{ {Figures/}}
\usepackage{epsfig} 
\usepackage{mathptmx} 
\usepackage{times} 
\usepackage{amsmath} 
\usepackage{amssymb}  
\usepackage{wasysym}
\usepackage{siunitx}
\usepackage{morefloats}
\usepackage[noadjust]{cite}

\usepackage{color,soul}
\usepackage{textcomp}
\usepackage{array}
\usepackage{makecell}
\usepackage{tabularx}
\usepackage{tabulary}

\title{\LARGE \bf
Design and Prototyping of a Bio-inspired Kinematic Sensing Suit for the Shoulder Joint: Precursor to a Multi-DoF Shoulder Exosuit
}
\author{Rejin John Varghese, Benny P L Lo, and Guang-Zhong Yang, \it Fellow IEEE
\thanks{*This research is supported by the Future AI and Robotics for Space (FAIR-SPACE) grant (EP/R026092/1) awarded by the EPSRC, UK.} 
\thanks{$^{*}$The authors are with The Hamlyn Centre, Department of Computing, Institute for Global Health Innovation, Imperial College London, SW7 2AZ, London, UK (email: 
	{\tt\small r.varghese15@imperial.ac.uk})}%
}

\begin{document}
\maketitle
\thispagestyle{empty}
\pagestyle{empty}

\begin{abstract}

Soft wearable robots are a promising new design paradigm for rehabilitation and active assistance applications. Their compliant nature makes them ideal for complex joints like the shoulder, but intuitive control of these robots require robust and compliant sensing mechanisms. In this work, we introduce the sensing framework for a multi-DoF shoulder exosuit capable of sensing the kinematics of the shoulder joint. The proposed tendon-based sensing system is inspired by the concept of muscle synergies, the body's sense of proprioception, and finds its basis in the organization of the muscles responsible for shoulder movements. A motion-capture-based evaluation of the developed sensing system showed conformance to the behaviour exhibited by the muscles that inspired its routing and validates the hypothesis of the tendon-routing to be extended to the actuation framework of the exosuit in the future. The mapping from multi-sensor space to joint space is a multivariate multiple regression problem and was derived using an Artificial Neural Network (ANN). The sensing framework was tested with a motion-tracking system and achieved performance with root mean square error (RMSE) of $\approx 5.43\,^{\circ}$ and $\approx 3.65\,^{\circ}$ for the azimuth and elevation joint angles, respectively, measured over 29,000 frames (4+ minutes) of motion-capture data.
\end{abstract}

\section{INTRODUCTION}
\label{sec:intro}

A significant fraction of the world population suffers from conditions affecting motor function. Additionally, an increasing average world population age will lead to increasing ageing-related debilitating muscular manifestations \cite{WorldHealthOrganization2011}. The increasing demand for the time-consuming, expensive and repetitive rehabilitation therapy and assistance traditionally provided by therapists and care workers is now making way for robot-based therapy and exoskeletons \cite{Varghese2018,Perry2007,Nef2009}. 
Rigid-bodied exoskeleton designs inherently have limitations like increased size/weight, lack of compliance, restrictive movements, and introducing misalignments \cite{Perry2007,Nef2009,Varghese2018}. The challenges faced in laboratory to real-world translation of rigid-bodied exoskeletons and the more dire need for systems that provide partial rather than complete assistance has resulted in a transition towards compliant, portable and wearable systems \cite{Varghese2018}. A new generation of soft wearable robots leveraging the body's anatomical structures is making use of unconventional materials and soft actuation methods for rehabilitation and assistance applications \cite{Varghese2018,xiloyannis2016modelling,Polygerinos2015a,Galiana2012, Lessard2017}. 

The compliant nature of soft exoskeletons (exosuits) makes them ideal for multiple degrees of freedom (DoF) joints like the shoulder that have inherent sources of misalignments. A few soft wearable robotic systems have focussed on 1-DoF (abduction/adduction) assistance for the shoulder \cite{Lessard2017,Galiana2012}. 2-DoF solutions include a semi-rigid continuum robot system \cite{Xu2013} and two multi-DoF tendon-driven exosuits \cite{Li2018, Gaponov2017}. 
Soft kinematic sensing networks are vital for compensating the compliance and non-linearities inherent in these human-exosuit systems, and needs to work alongside a vision-based system to provide sensory feedback for accomplishing reaching and manipulation tasks. This is analogous to the human body where the sense of proprioception and vision work together for natural and intuitive upper-limb functionality. Research works have explored different soft kinematics sensing systems including dielectric elastomers \cite{Xu2013a}, micro-fluidics \cite{Chossat2015}, and liquid metal alloys \cite{Menguc2014},  along with traditional sensors like flex/bend sensors \cite{Galiana2012} and Inertial Measurement Units (IMUs) \cite{Galiana2012}. Most of these systems suffer from increased hysteresis, limited range and other non-linearities. IMUs can achieve high performance, but suffer from drift \cite{Galiana2012,Aslani2018}. IMUs were not considered as our system is also being developed to counteract the muscular atrophy experienced by astronauts during prolonged space travel \cite{fitts2001functional}. IMUs need stable magnetic and gravitational fields to sense joint kinematics reliably.


\begin{figure*}[t]
\centering
\includegraphics[width=0.9\linewidth]{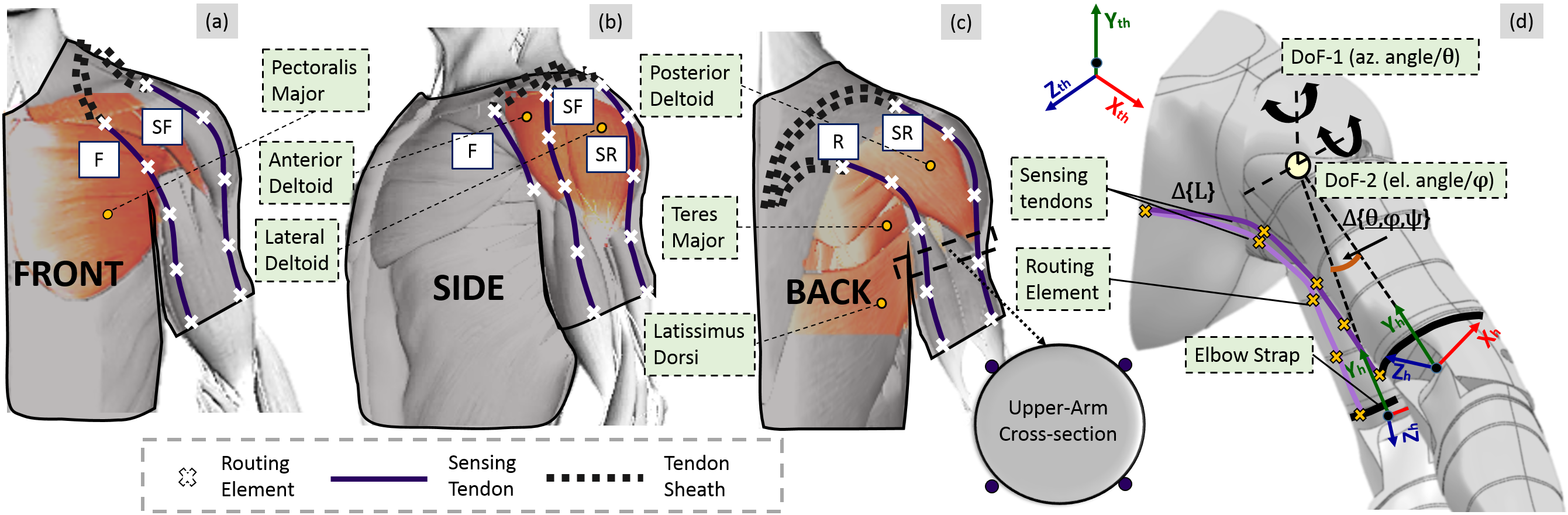}
\caption{Schematic of the sensing framework and principle: a)-c). Layout of the routing of the different tendons (tendons \textbf{F, SF, SR, R}) based on the different sets of muscles on the left upper-arm. The tendons are routed through 3-D printed elements and tendon-sheaths. A cross-section of the arm with the tendons is also shown. The figures are modifications of the schematic presented in \cite{Varghese2019BSN} d). Schematic of the sensing principle on a CAD rendering of the shoulder. The figure shows the effect of a change in the pose on the change in the lengths of a tendon routed across the shoulder.
}
\label{fig:ConceptSchematic}
\end{figure*}


In this work, we introduce the sensing framework of our multi-DoF shoulder exosuit which is based on a novel bio-inspired tendon-routing design architecture. The tendon-routing is inspired by the concept of muscle synergies \cite{Budhota2017}, and the sensing modality in the suit tries to replicate the mechanism behind our sense of proprioception \cite{Proske2009}. 
A preliminary simulation-based feasibility study of this framework was presented by the authors in \cite{Varghese2019BSN}. Tendon-driven systems, like the muscles in our body, can only apply forces while pulling and hence their positioning/routing on the body is crucial. There are infinite possible combinations of tendon-routing and, this study, therefore, is also a precursory feasibility study for the actuation framework of our exosuit.
In the next section, we introduce the bio-inspired tendon routing concept and design of our proposed sensing framework.


\section{SENSING FRAMEWORK DESIGN}
\label{sec:sysDes}

\subsection{The Bio-inspired Tendon-Routing Architecture}
\label{subsec:softSuit}

Traditionally, kinematic sensing in a robot is achieved through joint encoders or IMUs, but these solutions cannot be applied to our exosuit. As the suit aims to augment the body, we take inspiration from the kinematic sensing in the body. Proprioception, the ``sixth sense", is the sense of relative position/movement, force/effort, and balance \cite{Proske2009}. Muscle spindles in the skeletal muscle act as ``stretch receptors" (measuring stretch in the muscles and not joint angles) to get sense of a limb's position and movement \cite{Proske2009,Winter2005}. 

\textit{We hypothesize that a sensing framework based on measuring the stretch/displacement in multiple tendons routed around the shoulder (inspired by the organization of muscles influencing shoulder movement) could function analogous to the body's sense of proprioception and provide the exosuit with kinematic sensing capability} (see Fig.\ref{fig:ConceptSchematic}).

The glenohumeral (shoulder) joint is a complex ball-and-socket joint with 3+ DoFs \cite{Terry2000}. In this work, we aim to sense the 2-DoFs (azimuth(az./$\theta$) \& elevation(el./$\phi$) angles) that manifest conjointly (see Fig.\ref{fig:ConceptSchematic}(d)). The third DoF (internal/external rotation) manifests at the forearm and can only be measured at the elbow. Flexion/Extension (F/E), Abduction/Adduction (Ab/Ad)  \& Horizontal Abduction/Adduction (HAb/HAd) movements are facilitated by 5 sets of muscles (see Fig.\ref{fig:ConceptSchematic}): Chest (Pectoralis major (PM)), Back (Latissimus dorsi (LDo), Teres Major (TM)) and the three deltoid (shoulder) heads-Anterior (AD), Lateral (LD) and Posterior (PD) \cite{Terry2000}. Other muscles such as the biceps, triceps, and brachialis muscles also support shoulder movements \cite{Terry2000,Budhota2017}. If the tendon routing in the suit is replicated exactly like the muscles around the shoulder we would need 5+ tendons for sensing the 2-DoFs making the suit bulky and complex.

To reduce the number of tendons, we take inspiration from the muscle synergy concept, proposed by Bernstein \cite{Bernstein1967}. It states that the Central Nervous System simplifies movement control by co-activating a group of muscles responsible for a particular movement as a synergy rather than activating muscles individually. Even though 5+ muscles work together to generate movements in 2-DoFs, 4 muscle synergies can account for these movements \cite{Budhota2017,d2013control}. An example of a robotic glove employing a  synergy-based design is presented in \cite{xiloyannis2016modelling}. We, therefore, attempt a similar dimensionality reduction by routing tendons parallel ($\parallel$) to a set of muscles that work together and propose a tendon-routing architecture made up of 4 tendon-based sensing units- $\textbf{F, SF, SR, R}$ (see Fig.\ref{fig:ConceptSchematic}): 
\begin{itemize}
\item Tendon $\textbf{F}$ ${\parallel}$ (PM and AD): for sensing F/E and HAb/HAd.
\item Tendon $\textbf{SF}$ ${\parallel}$ (AD and LD): for sensing F/E and Ab/Ad.
\item Tendon $\textbf{SR}$ ${\parallel}$ (LD and PD): for sensing F/E and Ab/Ad.
\item Tendon $\textbf{R}$ ${\parallel}$ (PD, TM and LDo): for sensing F/E and HAb/HAd.
\end{itemize}

The tendons are routed along specific paths on the suit using routing elements (see Fig.\ref{fig:ConceptSchematic}\&\ref{fig:hardwareModel}). When the arm moves, the tendon paths (just like the associated muscle spindles in the muscles) shorten/elongate and this displacement/stretch is tracked by sensors (see Fig.\ref{fig:ConceptSchematic}(d)). Data from multiple tendons are then fused to derive the joint angles. The sensing framework w.r.t the shoulder's forward kinematics can be expressed with the following equations. The shoulder surface as a function of $(\theta,\phi)$ can be expressed parametrically as:
\begin{equation}
\label{eq:parametric}
\begin{aligned}
x_{S} = f_{\theta,\phi}(t,s), y_{S} = g_{\theta,\phi}(t,s), z_{S} = h_{\theta,\phi}(t,s)
\end{aligned}
\end{equation}
All routing elements will satisfy eq.\ref{eq:parametric}. The path traversed by the tendons can be expressed parametrically as:
\begin{equation}
\label{eq:tendonpath}
x_{t}= p(u), \text{  } y_{t} = q(u), \text{  } z_{t} = r(u)
\end{equation}
, and eq.\ref{eq:tendonpath} will satisfy eq.\ref{eq:parametric} at every point as the tendons lie on the shoulder surface at all times. The path length of the tendons can then be written as a line integral:
\begin{equation}
\label{eq:arcLen}
\ell_{t} = \int_{t}(x_{S},y_{S},z_{S})ds 
\end{equation}
, where $ds$ is an incremental step traversed over the length of the tendon. It can also be written as
\begin{equation}
\label{eq:arcLen2}
\begin{gathered}
\ell_{t} = \int{d\ell} \text{,where } d\ell = \sqrt[]{{dx_{t}}^2 + {dy_{t}}^2 + {dz_{t}}^2}
\end{gathered}
\end{equation}
The length of a tendon as a function of $(\theta,\phi)$ can be written as eq.\ref{eq:deltaarcLen}, and the sensing framework estimates ${\Delta}\ell(\theta,\phi)$  
\begin{equation}
\label{eq:deltaarcLen}
\begin{gathered}
\ell = L(\theta,\phi) \\
{\Delta}\ell(\theta,\phi) = L(\theta,\phi) - L({\theta}_{init},{\phi}_{init})
\end{gathered}
\end{equation}
, where (${\theta}_{init}$,${\phi}_{init}$) are the angles when the arm is in the rest position (neutral) by the side of the body. ${\Delta}\ell(\theta,\phi)$ is the metric estimated by each sensor. ${\textbf{S}} =   \{\Delta{\ell}_{F},\Delta{\ell}_{SF},\Delta{\ell}_{SR},\Delta{\ell}_{R}\} \subset \Re^4$ represents the sensor space made up of the 4 sensors, and $\textbf{q} = \{\theta,\phi\} \subset \Re^2$ represents the joint-space made up of azimuth and elevation joint angles. In the next section, we discuss the $\textbf{S}\mapsto\textbf{q}$ mapping to estimate the two joint angles (${\theta}$, ${\phi}$) of the shoulder joint from the four sensors.

\subsection{Sensor-Space to Joint-Space Mapping}
\label{subsec:sensorMapping}
To obtain joint angles from real/virtual sensors values, the mapping from sensor-space to joint-space ($\textbf{S}\mapsto\textbf{q}$) needs to be derived. While performing shoulder movement experiments, the joint angles (from motion capture data) and the corresponding sensor values are recorded and used to derive this ($\textbf{S}\mapsto\textbf{q}$) mapping. This $\textbf{S}\mapsto\textbf{q}$ mapping is equivalent to the inverse kinematics solution of the upper-arm in tendon-path space, where the termination point of each tendon can be considered the tip of the end-effector. This mapping from four sensor to two joint values is a multivariate multiple regression problem and can been solved very effectively by ANNs \cite{Varghese2019BSN}. The ANN behaves as a sensor fusion mechanism to fuse data from 4 homologous sensors to obtain 2 joint angles. We have earlier demonstrated this previously in our simulation-based work in \cite{Varghese2019BSN}. A similar framework is used here to extract the joint angles from the subject wearing the suit. Mathematically, this ANN-based $\textbf{S}\mapsto\textbf{q}$ mapping can be written as:
\begin{equation}
\label{eq:NNfiteq}
\textbf{q} = A(W\textbf{S} + b)
\end{equation}
, where $A$ is the activation function, $W$ the matrix of learned weights, and $b$ the bias values array.

\subsection{Hardware Design of the Sensing Suit}
\label{subsec:sensingHardware}
\begin{figure}
\centering
\includegraphics[width = 0.9\linewidth]{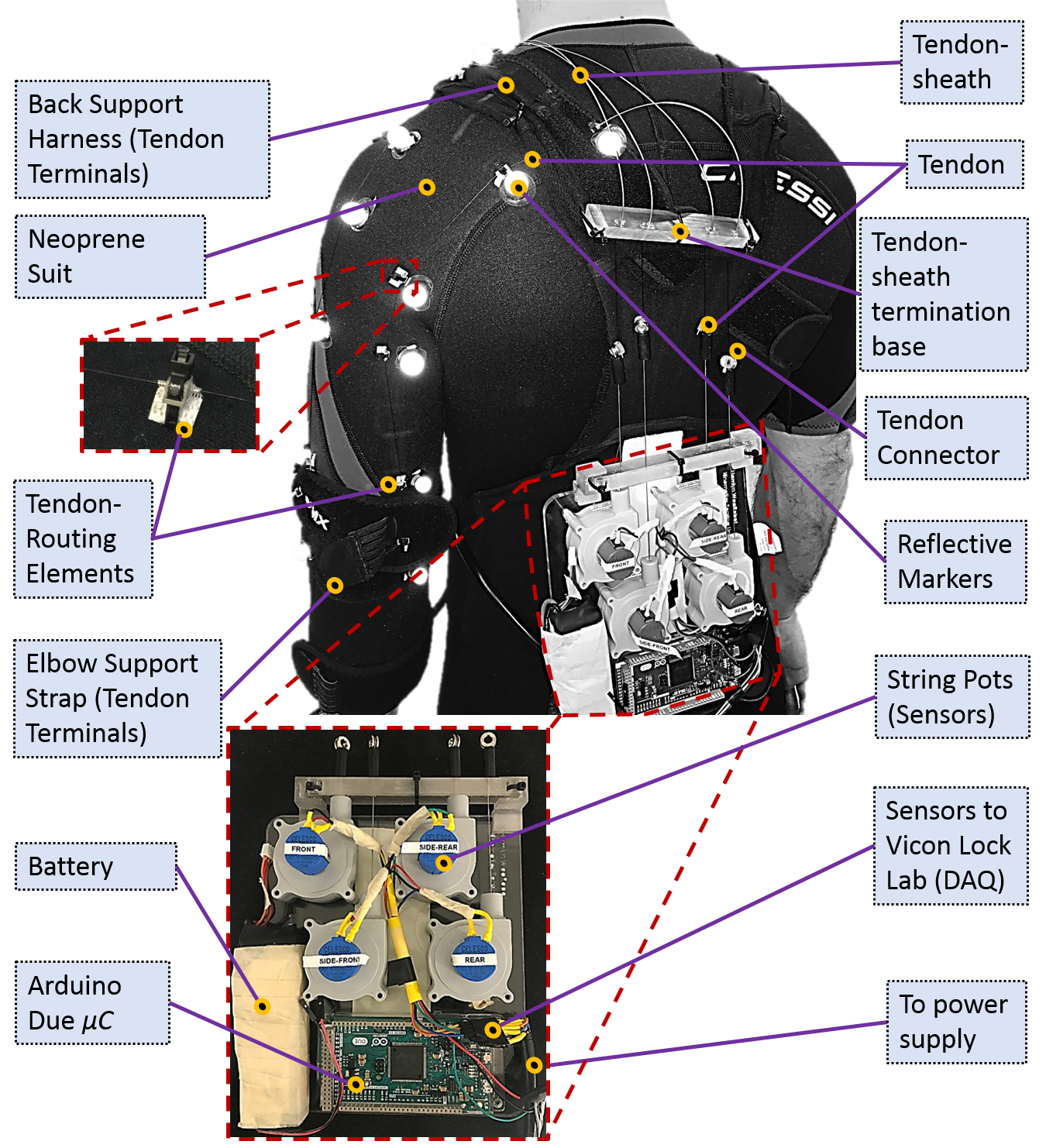}
\caption{Overview of the sensing framework prototype for the shoulder exosuit with a description of the electro-mechanical and mechanical elements.}
\label{fig:hardwareModel}
\end{figure}

To test our hypothesis, the suit presented in Fig.\ref{fig:hardwareModel} was developed personalized to the subject's physiology. A neoprene compression-fit suit (Cressi, Italy) was chosen as the base layer of the suit as it conforms closely with the skin and can withstand small forces from the tendons. The SUS 304 stainless steel tendons (Asahi Intecc Co. Ltd., Japan) were routed along the lines as discussed previously (see Fig.\ref{fig:ConceptSchematic}). The tendons are routed using 3D-printed routing elements attached to the base layer, and tendon-sheaths (Asahi Intecc Co. Ltd., Japan) in a Bowden-cable arrangement. 

The metal routing elements (see inset in Fig.\ref{fig:hardwareModel}) were printed using the Mlab Cusing 3D printer (Concept Laser, Germany). These routing elements ensure that the tendons maintain the desired path and adhere closely to the suit, and not follow the shortest line outside the body as seen in \cite{Li2018,Gaponov2017}. Both tendons (OD $\diameter0.47\si{\milli\meter}$) and the tendon-sheaths (OD $\diameter1.12\si{\milli\meter}$, ID $\diameter0.55\si{\milli\meter}$) were PTFE-coated to reduce friction. Small pieces of tendon-sheaths were also glued to the routing elements to reduce friction and allow smooth transition across the element. Between 4 and 6 elements were used to route each sensing tendon. A wearable back-support and elbow support sleeve (both fastened to the base) were used to support the routing elements that experienced largest forces (tendon termination points and where they exit out of the tendon sheaths). The tendon sheaths were used to route the tendons from the last routing elements to a 3D-printed termination base. This leaves the tendon exposed between the tendon-sheath termination to the origin of the sensor which is a string potentiometer (string-pot) as can be seen in Fig.\ref{fig:hardwareModel}. The tendons from the string-pots are then connected to the tendons across the shoulder using detachable 3D-printed parts. This was intentionally done for visual demonstrability of the concept (see accompanying video) and to make the electronics modular and separable. 

 \begin{figure*}[t]
\centering
\includegraphics[width = 0.9\linewidth]{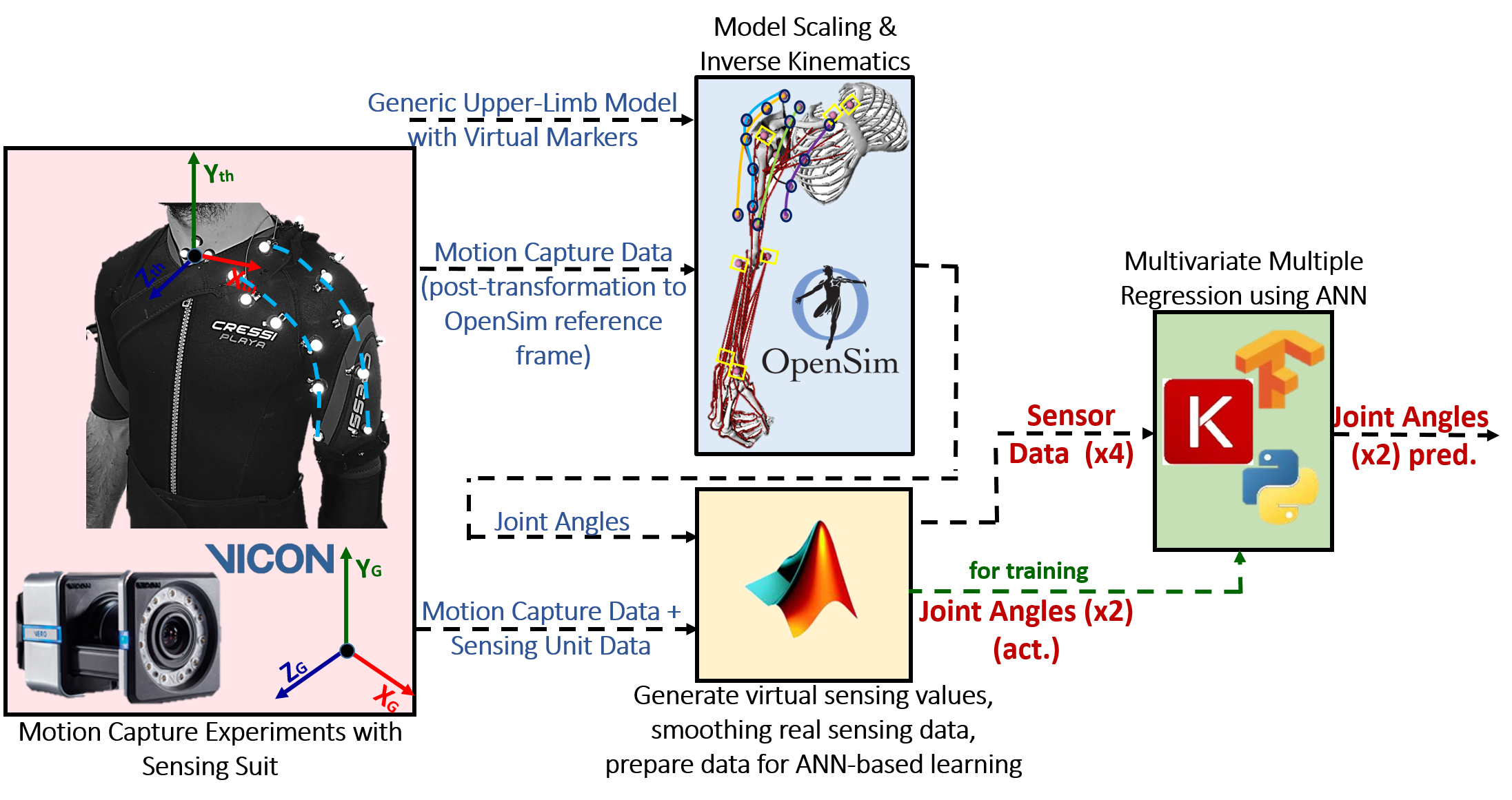}
\caption{Overview of the experimental methodology for prototype evaluation: Data from motion capture experiments are processed using both the OpenSim MS modelling software and in MATLAB to get the joint angle data and the real and virtual sensor data. This data is used to train an ANN to predict the joint angles given the 4 sensor values. Virtual sensor lines generated using reflective markers corresponding to the routing elements are shown through the dashed cyan-coloured lines.}
\label{fig:sysEvalOverview}
\end{figure*}

To measure the displacement of each tendon, its free end is attached to a Celesco SP1-12 string-pot sensor (Intertechnology Inc., Toronto, Canada). The spring in the sensor applies a constant small force ($\sim1.9\si{\newton}$) over a displacement of $12" (30.48\si{\centi\meter})$. This allows the tendon to be tracked accurately while being small enough to not deform the neoprene and fabric reinforcements excessively. The string-pot uses a potentiometer (10k\si{\ohm}) to measure the displacement through a voltage divider circuit. The analog signals from the four sensors are read by an Arduino Due (Arduino, Italy) microcontroller ($\mu$C) board. The input terminals of the string-pot were powered with $3.3\si{\volt}$, and the ADC on the Arduino $\mu$C was set to 12 bits to give the string-pot a theoretical sensor resolution of $\approx0.075\si{\milli\meter}$. 
The electronics were mounted on the back support making the entire system portable (see Fig.\ref{fig:hardwareModel}). The total weight of the electromechanical components and the base is $\sim950\si{\gram}$.




\section{CONCEPT EVALUATION EXPERIMENTS}
\label{sec:sysEval}
\subsection{Experimental Setup}
\label{subsec:expSetup}
To evaluate our prototype, experiments were performed with motion capture (MoCap) equipment to obtain ground truth data. The experiments were performed under the ethics approval provided by the Imperial College Research Ethics Committee with ICREC reference number 18IC4816. We used a lab with 10 Vero 2.2 cameras, a Vue video camera, a Lock Lab and auxiliary equipment from Vicon (Vicon, Oxford, UK) for our experiments. The subject was made to wear the prototype and the reflective markers and electronics were attached. 
For the MoCap experiments, the sensors were connected directly to the Vicon Lock Lab, a connectivity device to acquire external analog signal data and synchronize it with the motion capture data. The Lock Lab was set to acquire data from the sensors at 1200Hz and the cameras at 120Hz. Calibration experiments were performed on the string-pots to estimate displacements from sensor voltages. 

\begin{figure*}
\centering
\includegraphics[width = 0.9\linewidth]{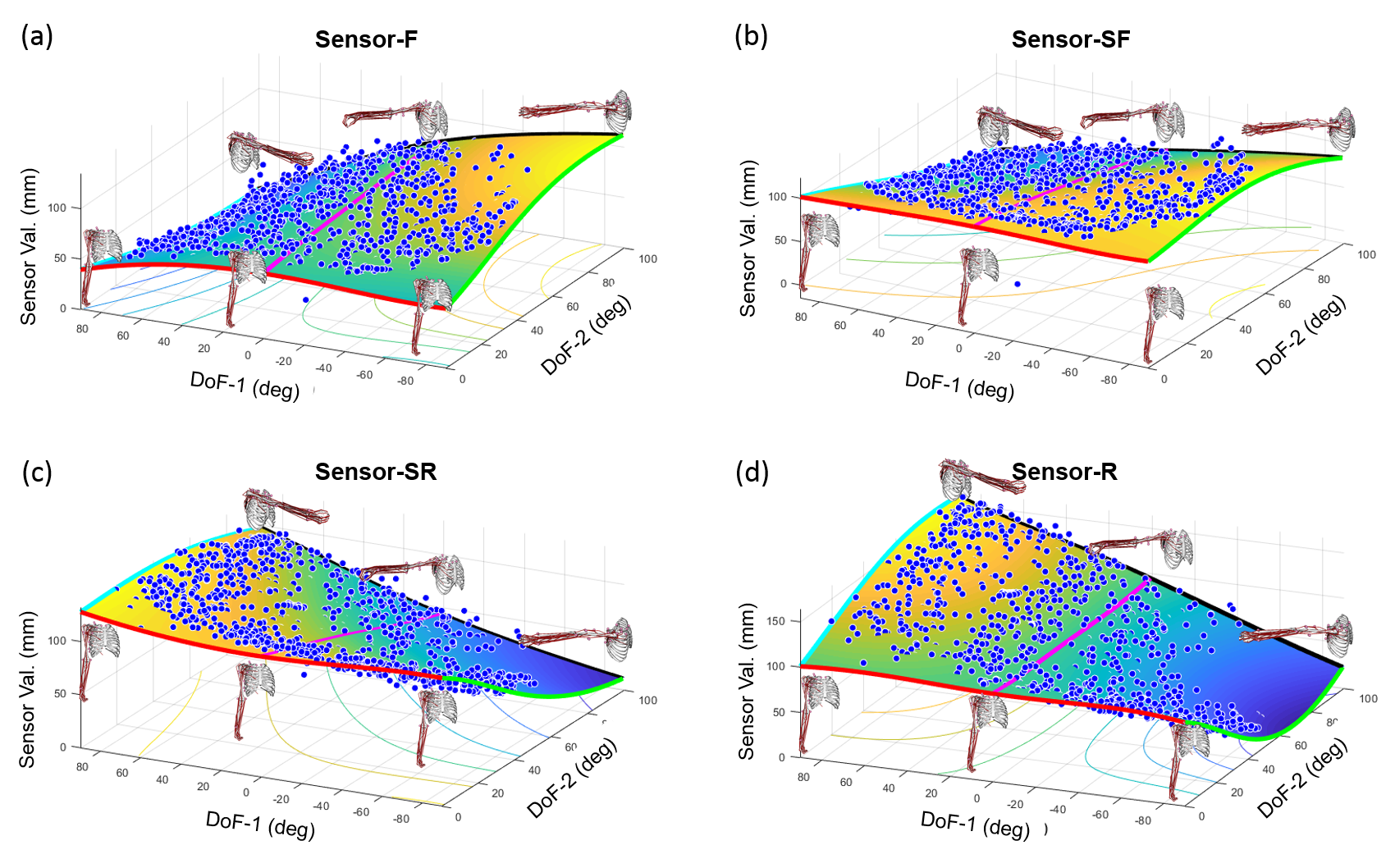}
\caption{$\textbf{q}\mapsto\textbf{S}$ Forward Kinematics Mapping: Change in path lengths measured by the 4 virtual sensors - a).\textbf{F}, b).\textbf{SF}, c).\textbf{SR} and , d).\textbf{R} over entire range of azimuth (DoF-1) and elevation (DoF-2) joint angles. The blue dots in the graph are $10\%$ of the sensor values, and the surface plot is the ANN-based $\textbf{q}\mapsto\textbf{S}$ fitting. In the graphs, the cyan and green lines represent F/E movements, the magenta line represents Ab/Ad movements, and the black line represents HAb/HAd movements. The red line represents the neutral position. The OpenSim graphics show the upper-limb position at the extremities of each line.}
\label{fig:arcLenvsAngles}
\end{figure*}

\subsection{Experimental Methodology}
\label{subsec:expMethod}
The experiments in this work were performed only on one subject as the tendon-routing architecture was designed specifically based on the physiology and morphology of that subject. To perform the motion-capture (MoCap) experiments, markers were attached adjacent to the 3D-printed routing elements and also on the subject's anatomical features like clavicle and elbow (see Fig.\ref{fig:hardwareModel}\&\ref{fig:sysEvalOverview}). The MoCap data from the anatomical features were used to derive the joint angles of the shoulder. OpenSim, an open-source musculoskeletal (MS) modelling software, was used \cite{Delp2007} to derive the same. We used the MoBL-ARMS Dynamic Upper Limb model developed  by  Saul \textit{et al.} in  \cite{Saul2015}  to solve the inverse kinematics of the shoulder. MATLAB (Mathworks, MA, USA) was used to transform the marker data from the Vicon system's reference frame to a local coordinate system on the subject used by OpenSim (see Fig.\ref{fig:sysEvalOverview}). Data from static experiments were then used to scale the generic MS model to our subject using OpenSim's Scaling tool. The OpenSim Inverse Kinematics tool was then applied on data from the dynamic MoCap experiments to derive the two DoFs. 
The movements performed during MoCap experiments for deriving and validating the $\textbf{S}\mapsto\textbf{q}$ map is presented in Table \ref{tab:exp_mvmts}. The table details the movements performed, joint angle ranges/limits during these movements, number of repetitions (reps) performed, and the trial time/frames of data recorded at 120Hz. A total of over 29,000 frames (4+ minutes) of data was obtained. Obtained joint angle data is based on the convention of International Society of Biomechanics \cite{Saul2015}.

\newcolumntype{K}[1]{>{\centering\arraybackslash}m{#1}}
\begin{table}
\renewcommand{\arraystretch}{1.2}
\caption{Movements Performed by Subject During MoCap Experiments}
\label{tab:exp_mvmts}
\centering
\begin{tabular}{K{1.95cm}||K{2cm}|K{0.8cm}|K{0.85cm}|K{0.8cm}}
\hline
\bfseries Movement & \bfseries Joint Angles Range/Limits & \bfseries No. of Reps. & \bfseries No. of Frames & \bfseries Trial Time \\
\hline\hline
\bfseries F/E & \makecell{$\theta={-90}^{0}/{90}^{0}$ \\ ${0}^{0}\leq\phi\leq{90}^0 $} & 4 & 3037 & 25.31s\\
 
\hline
\bfseries Ab/Ad & \makecell{$\theta={0}^{0}$ \\ ${0}^{0}\leq\phi\leq{90}^0$} & 4 & 3630 & 30.25s \\
 
\hline
\bfseries \makecell{Az. angle (fix.) \\ El. angle (var.)} & \makecell{$\theta=5$ const. vals. \\ ${0}^{0}\leq\phi\leq{90}^0$} & 2 & 5814 & 48.45s \\
\hline
\bfseries \makecell{El. angle (fix.) \\ Az. angle (var.)} & \makecell{ ${-40}^{0}\leq\theta\leq{90}^0$ \\ $\phi=5$ const. vals.} & 2 & 6757 & 56.31s\\
\hline
\bfseries Random movements & \makecell{${-40}^{0}\leq\theta\leq{90}^0$ \\ ${0}^{0}\leq\phi\leq{90}^0$} & N/A & 10313 &  85.94s\\
\hline
\multicolumn{3}{c}{\textbf{Total Frames/Total Time of Mocap Expts.}} & 29551 & 246.26s \\
\hline
\end{tabular}
\end{table}

The routing elements govern the behaviour of each sensing tendon (see eq.\ref{eq:parametric}-\ref{eq:deltaarcLen}), and hence were also tracked in the MoCap experiments. Along with the real sensor data, virtual sensors were derived using splines defined by joining these markers (see dashed cyan lines in Fig.\ref{fig:sysEvalOverview}). Splines were derived for each sensor for each frame of the captured data. The changes in lengths of these splines were computed using eq. \ref{eq:arcLen2}-\ref{eq:deltaarcLen}. The complete analysis of this framework using virtual sensing lines was also performed in MATLAB. A comparison between real and virtual sensor data is presented in Section \ref{subsec:realVsVirtRes}.

The virtual sensor and joint angle data from OpenSim were used to derive the $\textbf{q}\mapsto\textbf{S}$ map (forward kinematics) and the $\textbf{S}\mapsto\textbf{q}$ (inverse kinematics) map. The $\textbf{q}\mapsto\textbf{S}$ map helps us analyze the behaviour of the tendon-routing architecture (see Fig.\ref{fig:arcLenvsAngles}) and is discussed in Section.\ref{subsec:fKinRes}. The forward mapping is also derived using a very shallow ANN network using 1 hidden layer with 8 nodes. The methodology to obtain the inverse mapping ($\textbf{S}\mapsto\textbf{q}$) remains the same as was used in \cite{Varghese2019BSN}. To prepare the data for neural network-based fitting, the data was scaled globally and shuffled. To train and test the ANN, data from 29,551 frames of MoCap data was used. The ANN was trained on 65\% of the data, validated on 15\% of the data, and tested on 20\% of previously unused data. The results of this mapping is discussed in Section \ref{subsec:NNMapRes}. The network was developed using the Keras-TensorFlow API in Python 3.5, on an HP Workstation with Intel i7-6700 CPU processor @3.4GHz with 16GB RAM.

\section{RESULTS AND DISCUSSION}
\label{sec:resAndDisc}

\subsection{Tendon-Routing Architecture Behaviour Analysis}
\label{subsec:fKinRes}

The behaviour of the tendon routing architecture conceptualized in Section \ref{sec:sysDes} is analyzed here. Fig.\ref{fig:arcLenvsAngles} presents the forward kinematics mapping ($\textbf{q}\mapsto\textbf{S}$) of the data from the 4 virtual sensors on moving the shoulder in 2 DoFs.  Fig.\ref{fig:arcLenvsAngles} shows $10\%$ of the sensor values (blue dots) and the ANN-fitted surface.  Analyzing this mapping is the first step to the actuation framework of the exosuit. This study is crucial as there are infinite tendon-routing combinations, and that even though the sensing tendons can sense during both extending/retracting, actuating tendons can only apply forces when pulling (monotonously decreasing). 

We can use the $\textbf{q}\mapsto\textbf{S}$ map (in Fig.\ref{fig:arcLenvsAngles}) to predict if the actuation tendons will displace monotonously and apply the required forces, if the same routing was used for actuation as well. The tendon-routing architecture analysis presented here was further required to isolate the source of the non-monotonous tendon displacement behaviour observed when the same analysis was performed on the simulation-based study presented in \cite{Varghese2019BSN}. We hypothesized two potential sources for the non-monotonicity: 1). a sub-optimal tendon-routing architecture, or 2). the limitation in OpenSim which only allowed for markers to be attached to the bones rather than on the skin surface, as would happen in reality. The hypothesis of the tendon-routing architecture being the error source is invalidated in this study.


From Fig.\ref{fig:arcLenvsAngles}, it can be seen that the sensors' behaviour match the behaviour of the equivalent muscles/synergies on which their paths are based. For \textit{e.g.}, it can be seen that during flexion, sensor value $\textbf{F}$ monotonously decreases and sensor value $\textbf{R}$ monotonously increases, and vice-versa during extension (see cyan and green line in Fig.\ref{fig:arcLenvsAngles}(a)\&(d)). This is consistent with the behaviour of the AD, PM, TM and LDo muscles on which tendons $\textbf{F}$ \& $\textbf{R}$ have been based. Similarly, during abduction, sensors-$\textbf{SF}$ and $\textbf{SR}$ monotonously decrease (see the magenta line in Fig.\ref{fig:arcLenvsAngles}(b) \& (c)). This is in keeping with the behaviour of the  PD, AD and LD muscles along which the tendons have been routed. The increase/decrease in sensor value in Fig.\ref{fig:arcLenvsAngles} is reflected in the change in colour on the surface plot. For the HAb/Ad movements as well, the observed sensor data from all 4 sensors conform to the behaviour of the muscles along which they were routed. The results of this movement are shown in the black lines in the 4 subplots of Fig.\ref{fig:arcLenvsAngles}. We can consider these results the validation of our proposed bio-inspired sensing concept. The red line corresponds to the arm in neutral/rest position.

Some inconsistencies in Fig.\ref{fig:arcLenvsAngles} are: 1). The red line in Fig.\ref{fig:arcLenvsAngles}(a) should have been a straight line since it represents the hand in neutral position. The curve is observed as it is an extrapolation by the ANN-fitting. The arm cannot reach $<10^{\circ}$ in that DoF due to the thickness of the suit, and the network, makes an estimation along that line in Fig.\ref{fig:arcLenvsAngles}(a). 2) Some non-monotonicity in the ANN-fitted surface is seen in the top-right corner of Fig.\ref{fig:arcLenvsAngles}(b)\&(d). This is again an extrapolation by the fitting, as that region is out of the shoulder's reachable workspace (blue dots absent).

\begin{figure}
\centering
\includegraphics[width = 1\linewidth]{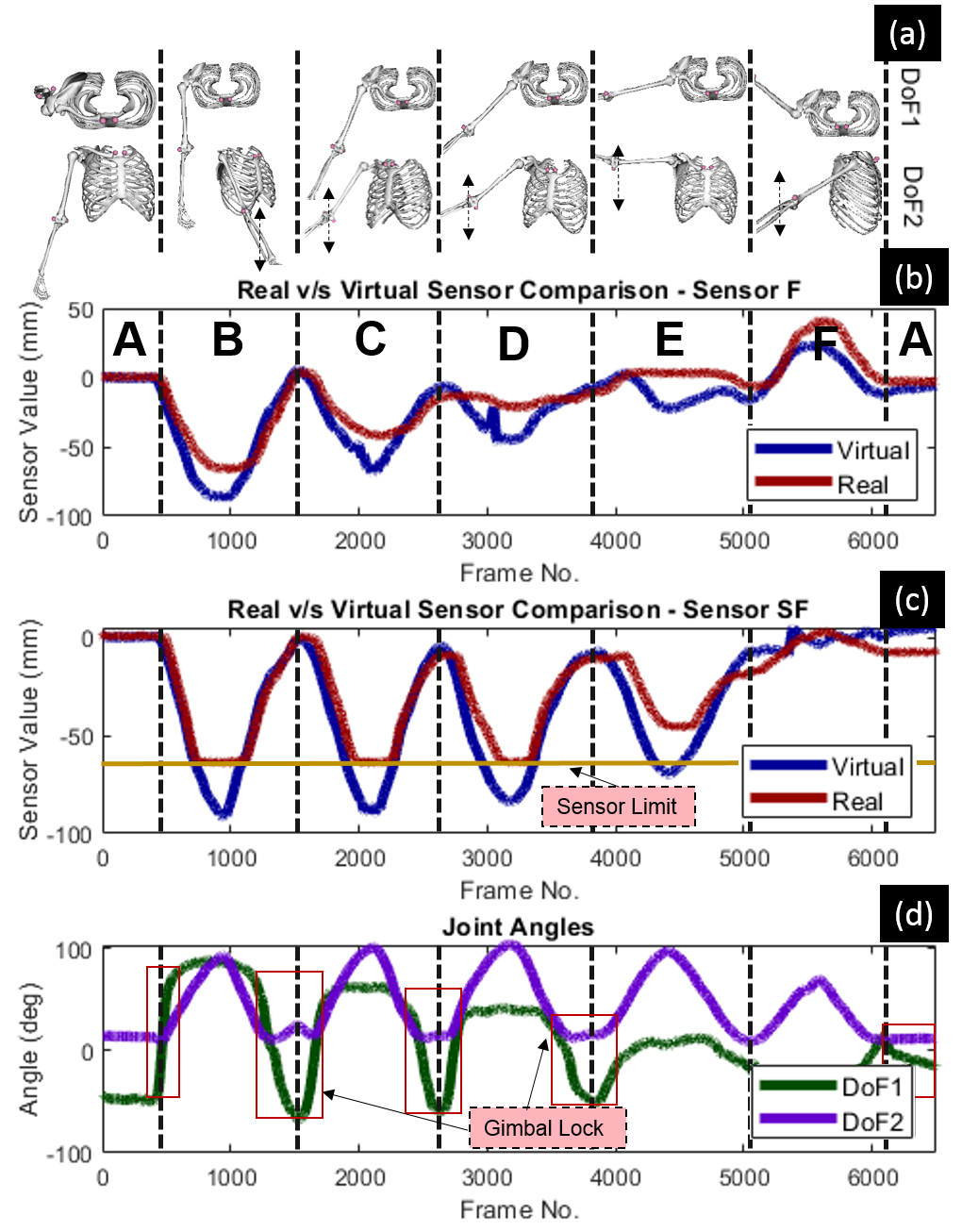}
\caption{Comparison of real and virtual sensor performance I: a). Graphics of the top and front view of the pose of the upper-limb during the trial. b)-c). Real and virtual sensor-\textbf{F} \& \textbf{SF} values obtained during the trial. d). DoF-1 \& DoF-2 values during the trial. DoF-1 is held constant at a value while DoF-2 is changed from neutral to parallel to ground configuration.}
\label{fig:realVirtCompare}
\end{figure}

\subsection{Real v/s Virtual Sensing Analysis}
\label{subsec:realVsVirtRes}
\begin{figure}
\centering
\includegraphics[width = 0.95\linewidth]{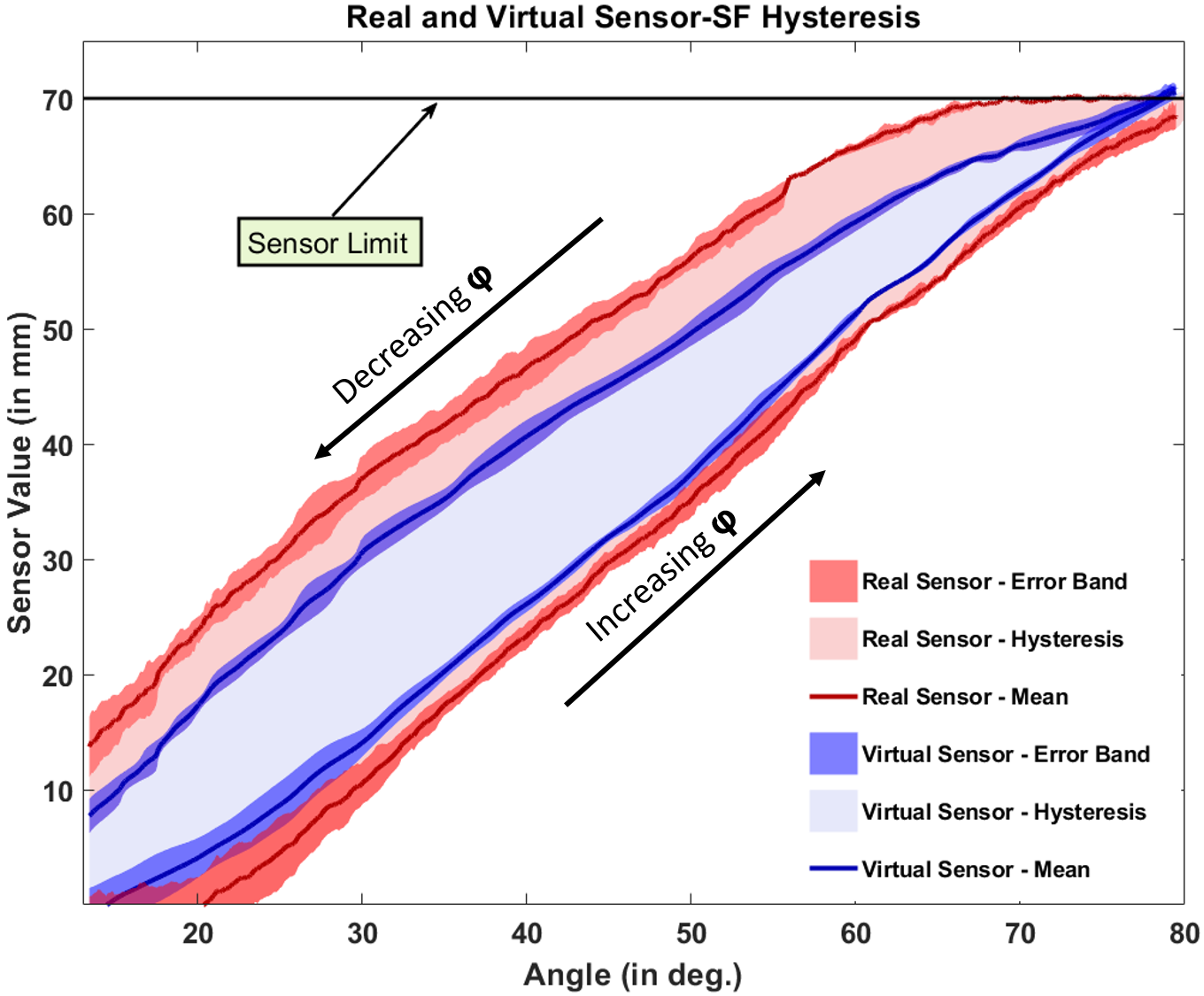}
\caption{Comparison of real and virtual sensor performance II: The graph shows the displacement of real (red) and virtual (blue) sensor-\textbf{SF} values against elevation angle during multiple F/E movements. This graph presents the hysteresis and error bands observed in both the real and virtual sensors.}
\label{fig:realVirtCompareII}
\end{figure}

To further study our sensing framework, virtual and real sensor values from a movement trial are compared in Fig. \ref{fig:realVirtCompare}\&\ref{fig:realVirtCompareII}. This analysis was done to understand the effects of compliance, friction and other non-linearities encountered by the real tendons vis-a-vis the unhindered virtual tendons. This sub-section presents the evolution of:
\begin{enumerate}
    \item real v/s virtual sensors w.r.t time (see Fig.\ref{fig:realVirtCompare}),
    \item real v/s virtual sensors w.r.t joint angles (see Fig.\ref{fig:realVirtCompareII}).
\end{enumerate}

  The OpenSim graphics in Fig.\ref{fig:realVirtCompare}(a) show the movement of the upper-limb in top and front/side view. The trial presented in Fig.\ref{fig:realVirtCompare} is detailed in Table \ref{tab:exp_mvmts} row 3 where the azimuth angle (DoF-1) is held at different constant values while varying the elevation angle (DoF-2). 5 movements (B-F) are performed in the trial where phase A, B and E represent neutral position, F/E and Ab/Ad respectively.

In Fig.\ref{fig:realVirtCompare}(b)\&(c), we observe that both real and virtual sensors follow the same trend lines for both sensors-\textbf{F}\&\textbf{SF}, and follow the behaviour of their associated muscles/synergies. For both sensors-\textbf{F}\&\textbf{SF} (see Fig.\ref{fig:realVirtCompare}(b)\&(c)), we see that the absolute peak value for both real and virtual sensors decrease with decreasing DoF-1 value. For sensor-\textbf{F}, the value goes almost to nil during Ab/Ad and changes direction when reverse F/E is performed, while for sensor-\textbf{SF}, the value follows the same trend as sensor-\textbf{F}, but goes to zero during reverse F/E movement. In sensor-\textbf{F}, there is some lag between the virtual and real sensor. This is observed because the tendon gets pinched near the axillary fossa (armpit) which increases friction and hinders movement. We intend to correct this in the next iteration by modifying the tendon-routing around the region. 

The joint angle representation used for this work is in spherical coordinate system, \textit{i.e.} the representation adopted by the MS model \cite{Saul2015}. This representation has been used instead of quaternions as it is more intuitive and affords better visualization of results. The downside of this representation is the Gimbal lock problem, and is seen in the sharp change in DoF-1 value as the arm moves to the neutral position (see Fig.\ref{fig:realVirtCompare}(d)). We intend to switch to quaternion representation when using this sensing framework for feedback control.

Fig.\ref{fig:realVirtCompareII} presents another comparison of the virtual and real sensor values when analyzed w.r.t joint angles. Fig.\ref{fig:realVirtCompareII} presents the evolution of sensor-\textbf{SF} value with changing elevation angle ($\phi$) during F/E movements. We can clearly see the existence of hysteresis behaviour in both the real and virtual sensors. In the case of virtual sensors, this behaviour is observed primarily due to the inherent soft nature of the neoprene base suit on which the markers are mounted. In the case of the real sensors, this behaviour is further exacerbated, because of the sensor limit imposed by the termination base (see Fig.\ref{fig:realVirtCompareII}). As previously mentioned, this was done for visual demonstrability of the concept, \textit{i.e.} to visualize tendon displacement as the shoulder moves (see accompanying video). The sensor is not able to recover completely from the imposed sensor limit as we have additional non-linearities introduced by: 1). the friction between tendons and the routing elements, tendon-sheaths, \textit{etc.}, and 2). the local pulling (compliance) experienced at the termination elements due to the spring force of the sensor transmitted by the tendon.  It is predominantly due to the imposed sensor limit that the tendon experiences a small permanent deformation at the end of sequence of F/E movements. Similar behaviour is observed in other movements as well. The small error band of observed real/virtual sensor values over repeated movements (see Fig.\ref{fig:realVirtCompareII}) demonstrates the repeatability of the framework. 
We intend to address the friction- and compliance-related shortcomings with adequate design countermeasures in the next version of the suit.     



\begin{figure}
\centering
\includegraphics[width = \linewidth]{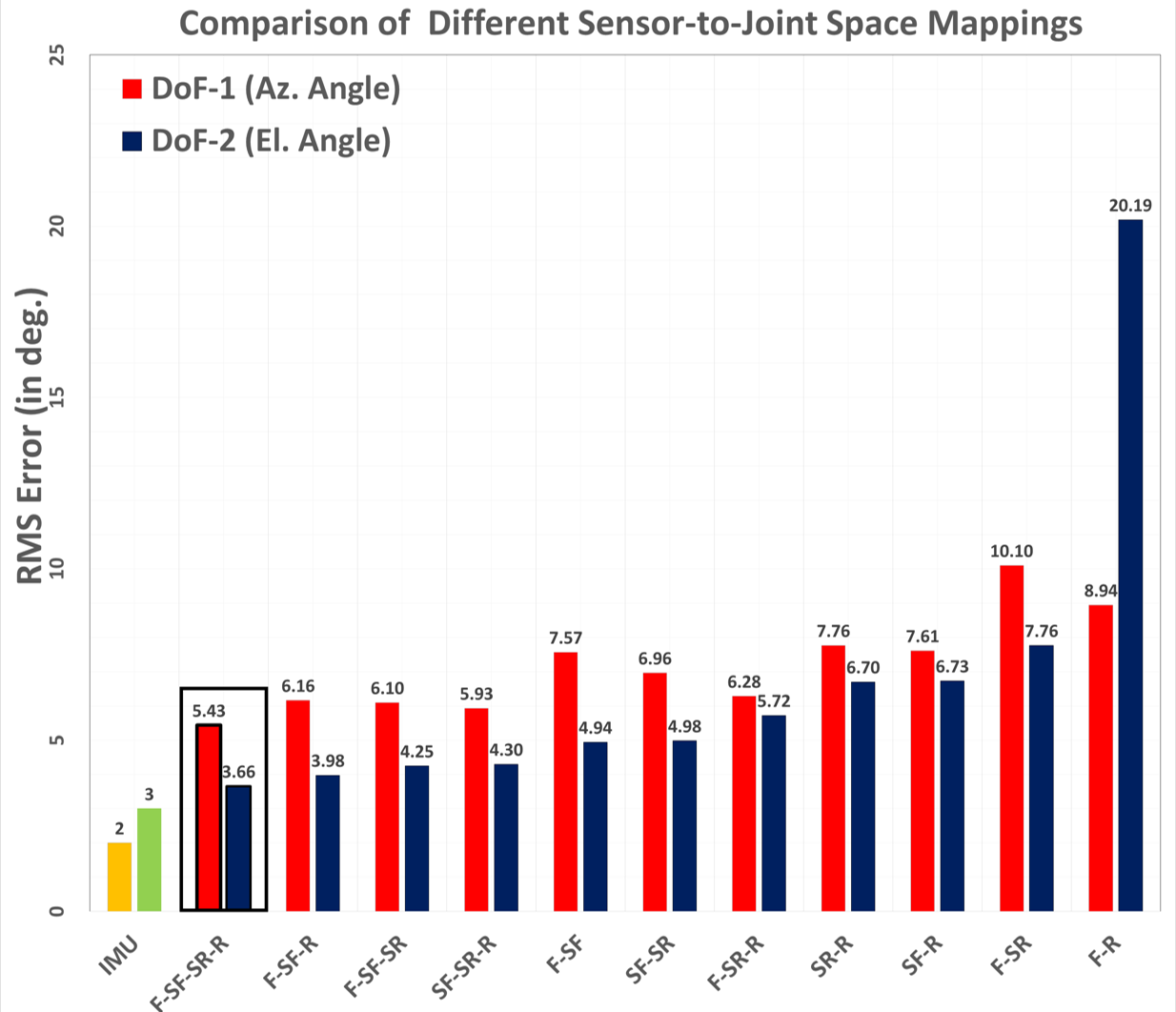}
\caption{$\textbf{S}\mapsto\textbf{q}$ Mapping Performance: RMSE for DoF-1 \& DoF-2 for an ANN with 1 hidden layer of 25 neurons. The sensor combination performing the best is encapsulated in the black box. The results for IMU-based solution is shown with orange (DoF-1) and green (DoF-2) bars.}
\label{fig:mapPerform}
\end{figure}

\subsection{Sensor-to-Joint Space Mapping Performance}
\label{subsec:NNMapRes}
This sub-section discusses the performance of the ANN-based $\textbf{S}\mapsto\textbf{q}$ mapping as presented in Fig.\ref{fig:mapPerform}.  Fig.\ref{fig:mapPerform} shows the performance of the $\textbf{S}\mapsto\textbf{q}$ mapping learned using different networks and varying number of virtual sensors as inputs to the network. The network using all the 4 sensors performed the best followed by comparable performances observed by different combinations of the three sensors. This is followed by the performances of different pairs of sensors. This behaviour is analogous to the behaviour observed in \cite{Budhota2017} where 4 muscle synergies accounted for more than 90\% of the variance in shoulder movements followed by comparable performance by 3 synergies and reduced performance by 2 synergies. In Fig.\ref{fig:mapPerform}, the anomaly is the combination of sensors-\textbf{F}\&\textbf{R}, which shows a significantly high RMS error. This is observed because sensors-\textbf{F}\&\textbf{R} are almost an agonist-antagonist pair, thereby bringing in very similar contributions to the $\textbf{S}\mapsto\textbf{q}$ map.
 We can see that an RMSE of $5.43^{\circ}$ in DoF-1 and $3.65^{\circ}$ in DoF-2 was achieved with a network having 25 neurons in the hidden layer and an input from all 4 sensors.  These results are promising as we have used a vanilla ANN-based fitting and have achieved RMSE comparable with state-of-the-art IMU-based sensing solutions in DoF-2 \cite{Aslani2018}. 
 
 As mentioned earlier, the advantage of our system over state-of-the-art IMU-based solutions is the ability to use the system in space environments as well, where IMUs could malfunction. From the size perspective, both systems can be very low-profile over the limbs and only require electronics that can be placed inside a backpack. Additionally, another advantage of this framework is the ability to use the sensing tendons as a benchmark for tendons in the actuation system for compensating backlash.  Due to the observed hysteresis (see Fig.\ref{fig:realVirtCompareII}), we hypothesize that the results of $\textbf{S}\mapsto\textbf{q}$ mapping can be further improved with history-based learning and statistical sensor-fusion methods like Recurrent Neural Networks (RNNs), Long Short-Term Memory (LSTM) networks and different Kalman filters.

\section{CONCLUSION}
\label{sec:conc}
In this work, we present the first step towards a low-profile and wearable multi-DoF shoulder exosuit. Through this paper, we strived to address the challenge of kinematic sensing of a complex joint like the shoulder where multiple DoFs manifest conjointly through our kinematic sensing framework. The novel tendon-routing architecture of our proposed  framework takes inspiration from the muscle synergies of the shoulder and our sense of proprioception. To the best of our knowledge, no such system has been developed. MoCap experiments were used to validate the performance of our tendon-routing architecture, and were also used to train an ANN-based mapping from sensor data to joint angles ($\textbf{S}\mapsto\textbf{q}$). The trained ANN with an input from the four sensors achieved a performance with an RMSE of $5.43^{\circ}$ and $3.65^{\circ}$ in estimating the azimuth and elevation joint angles, respectively.
	
The prototype presented here proved the feasibility of our proposed bio-inspired tendon-routing architecture. Based on this work, we think that this same tendon-routing architecture could be extended to other multi-DoF joints in the body like the hip, wrist and ankle. The work presented here is a precursor to extending the architecture to the actuation framework of the exosuit, as it validates the choice of tendon-routing compared to the infinite tendon-routing combinations that are possible. The suit was developed based on the physiology and morphology of one subject, and hence was analyzed on the same subject to validate our proposed bio-inspired tendon-routing architecture. We intend to develop more subject-specific prototypes to extend the study to multiple subjects. We are also working to address a few design-based shortcomings such as adding friction- and compliance-management elements to the suit. We are also exploring more sophisticated neural networks like LSTMs to better estimate shoulder joint kinematics. In the future, we intend to extend this concept to sense internal/external rotation of the shoulder as well.





\section*{ACKNOWLEDGMENT}
The authors would like to acknowledge the EPSRC, UK for funding this work. The authors would also like to thank Dr. Dennis Kundrat, Pierre Berthet-Rayne, Dr. Fani Deligianni, Xu Chen and Mohamed Abdelaziz for their support with the project, and Dr. Yao Guo, Daniel Freer, Ya-Yen Tsai and Daniel Bautista for their help with MoCap experiments.

\bibliographystyle{IEEEtran}
\bibliography{IEEEabrv,ICRA2019_SensingFramework_References}

\end{document}